\documentclass[10pt,twocolumn,letterpaper]{article}

\usepackage{iccv}
\usepackage{times}
\usepackage{epsfig}
\usepackage{graphicx}
\usepackage{amsmath}
\usepackage{amssymb}

\usepackage{bm}
\usepackage{bbm}
\usepackage{array}
\usepackage{pifont}
\usepackage{mathrsfs}
\usepackage{url}
\usepackage{multirow}
\usepackage{tablefootnote}
\newcommand{\tabincell}[2]{\begin{tabular}{@{}#1@{}}#2\end{tabular}} 
\usepackage{color}
\usepackage[ruled,linesnumbered]{algorithm2e}
\SetKwInput{KwInput}{Input}                
\SetKwInput{KwOutput}{Output}

\DeclareMathOperator*{\argmax}{arg\,max} 
\usepackage[pagebackref=true,breaklinks=true,letterpaper=true,colorlinks,bookmarks=false]{hyperref}

\iccvfinalcopy 

\def\iccvPaperID{6099} 

\begin{document}

\title{Towards Face Encryption by Generating Adversarial Identity Masks}

\author{Xiao Yang$^{1}$, Yinpeng Dong$^{1,3}$, Tianyu Pang$^{1}$, Hang Su$^{1,2}$\thanks{Corresponding author}, Jun Zhu$^{1,2,3}$, Yuefeng Chen$^{4}$, Hui Xue$^{4}$\\
$^{1}$ Dept. of Comp. Sci. and Tech., Institute for AI, BNRist Center, Tsinghua-Bosch Joint ML Center\\
$^{1}$ THBI Lab, Tsinghua University, Beijing, 100084, China \\
$^{2}$ Pazhou Lab, Guangzhou, 510330, China \hspace{2ex} $^{3}$ RealAI \hspace{2ex} $^{4}$ Alibaba Group\\
\tt\small{\{yangxiao19, dyp17, pty17\}@mails.tsinghua.edu.cn} \\ \tt\small{\{suhangss, dcszj\}@mail.tsinghua.edu.cn} \hspace{2ex} \{yuefeng.chenyf, hui.xueh\}@alibaba-inc.com}

\maketitle

\begin{abstract}
   As billions of personal data being shared through social media and network, the data privacy and security have drawn an increasing attention. Several attempts have been made to alleviate the leakage of identity information from face photos, with the aid of, e.g., image obfuscation techniques. However, most of the present results are either perceptually unsatisfactory or ineffective against face recognition systems.
    Our goal in this paper is to develop a technique that can encrypt the personal photos such that they can protect users from unauthorized face recognition systems but remain visually identical to the original version for human beings. 
    To achieve this, we propose a targeted identity-protection iterative method (TIP-IM) to generate adversarial identity masks which can be overlaid on facial images, such that the original identities can be concealed without sacrificing the visual quality. 
    Extensive experiments demonstrate that TIP-IM provides 95\%+ protection success rate against various state-of-the-art face recognition models under practical test scenarios. Besides, we also show the practical and effective applicability of our method on a commercial API service.
\end{abstract}



\section{Introduction}

The blooming development of social media and network has brought a huge amount of personal data (e.g., photos) shared publicly.
With the growing ubiquity of deep neural networks, these techniques dramatically improve the capabilities for the face recognition systems to deal with personal data~\cite{deng2019arcface,liu2017sphereface,schroff2015facenet,wang2018cosface}, but
as a byproduct, also increase the potential risks for privacy leakage of personal information.
For example, an unauthorized third party may scrabble and identify the shared photos on social media (e.g., Twitter, Facebook, LinkedIn, \emph{etc.}) without the permission of their owners, resulting in cybercasing~\cite{larson2018pixel}.
Therefore, it is imperative to provide users an effective way to protect their private information from being unconsciously identified and exposed by the excessive unauthorized systems, without affecting users' experience.


\begin{figure}[t]
\centering
\includegraphics[width=.95\linewidth]{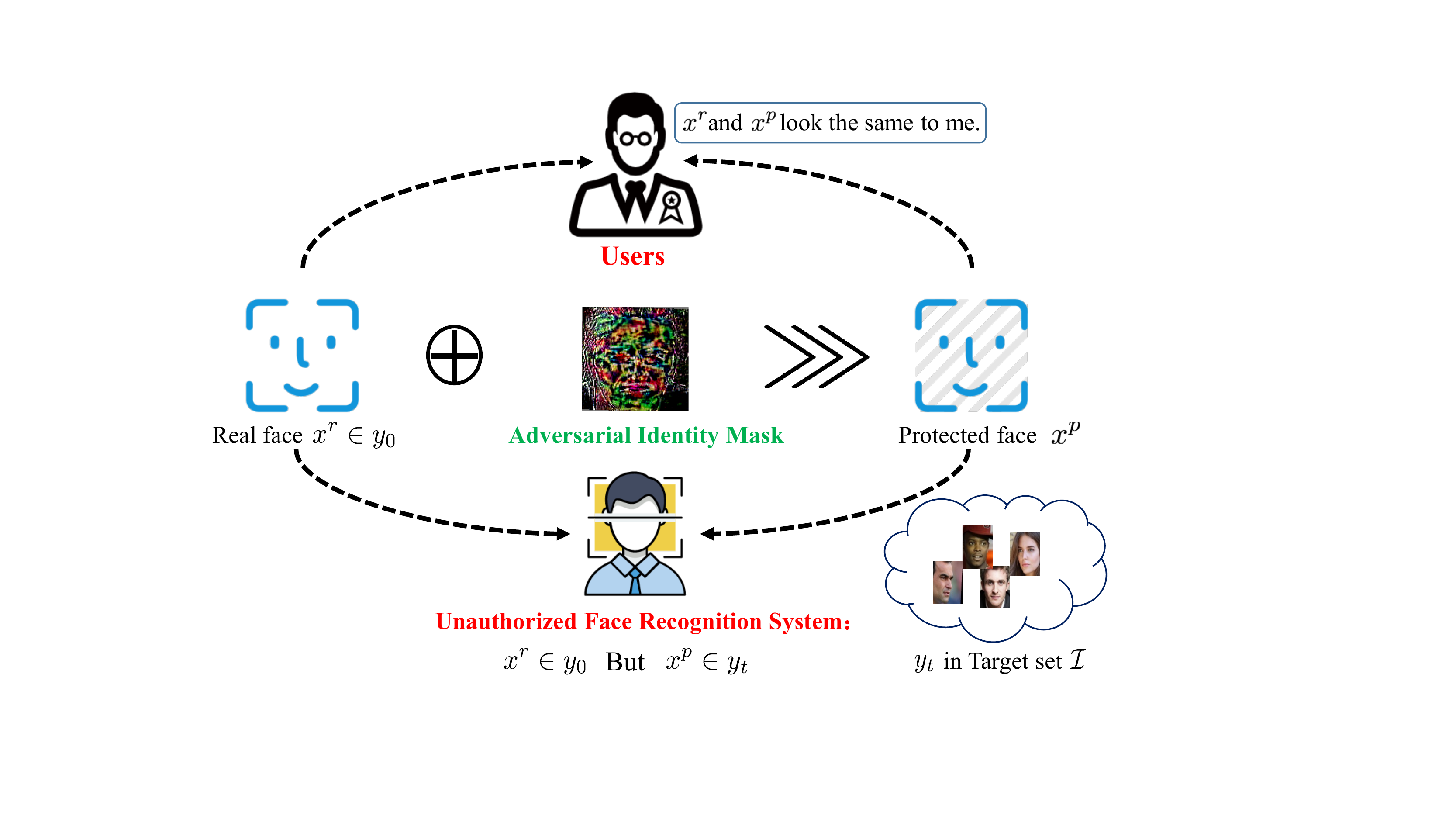}
\vspace{-1ex}
\caption{An illustrative example of targeted identity protection. When users share a photo $x^{r}$ on social media (e.g., Twitter, Facebook, \emph{etc.}), unauthorized applications could scrabble this identity $y_{0}$ based on face recognition systems, resulting in the privacy leakage of personal information. Thus we provide an effective identity mask tool to generate a protected image $x^{p}$, which can conceal the corresponding identity by misleading the malicious systems to predict it as a wrong target identity $y_{t}$ in an authorized or virtual target set, which can be provided by the service providers.  
}
\label{fig:intro}
\vspace{-3ex}
\end{figure}

The past years have witnessed the progress for face encryption in both the security and computer vision communities.
Among the existing techniques, obfuscation-based methods are widely studied. Conventional obfuscation techniques~\cite{wilber2016can}, such as blurring, pixelation, darkening, and occlusion, are maybe either perceptually satisfactory or effective against recognition systems~\cite{mcpherson2016defeating,oh2016faceless,rodrigues2006face}. The recent advance in generative adversarial networks (GANs)~\cite{goodfellow2014generative} provides an appealing way to generate more realistic images for obfuscation~\cite{gafni2019live,sun2018natural,sun2018hybrid,wu2018privacy,li2019anonymousnet}. However, the resultant obfuscated images have significantly different visual appearances compared with the original images due to the exaggeration and suppression of some discriminative features, and occasionally generate unnatural output images with undesirable artifacts~\cite{sun2018hybrid}. 



Recent researches have found that adversarial examples can evade the recognition of a  
FR system~\cite{yang2020delving,szegedy2013intriguing,goodfellow2014explaining,Sharif2016Accessorize} by overlaying adversarial perturbations on the original images~\cite{Athalye2018Obfuscated}.
It becomes an appealing way to 
apply an adversarial perturbation to conceal one's identity, even under a more strict constraint of impersonating some authorized or generated face images when available (e.g., given by the social media services). It provides a possible solution to specify the output, which may avoid an invasion of privacy to other persons if the resultant image is recognized as an arbitrary identity~\footnote{The practical FR system will obtain the similarity rankings from the candidate image library, which could be mistaken for someone else.}.
It should nevertheless be noted that although the adversarial perturbations generated by the existing methods (e.g., PGD~\cite{Kurakin2016} and MIM~\cite{Dong2017}) have a small intensity change (e.g., $12$ or $16$ for each pixel in $[0,255]$), they may still sacrifice the visual quality for human perception due to the artifacts as illustrated in Fig.~\ref{fig:perturbation}, and similar observation is also elaborately presented in~\cite{zhao2017generating,sen2019should} that $\ell_{p}$-norm adversarial perturbations can not fit human perception well.
Moreover, the current adversarial attacks are mainly dependent on   
either the \emph{white-box} control of the target system~\cite{Sharif2016Accessorize,oh2017adversarial} or the tremendous number of \emph{model queries}~\cite{dong2019efficient}, which are impractical in real-world scenarios (e.g., unauthorized face recognition systems on social media) for identity protection.



In this paper, we involve some valuable considerations from a general user's perspective and propose 
to alleviate the identity leakage of personal photos in real-world social media. 
We focus on face identification in particular, a typical sub-task in face recognition, the goal of which is to identify a real face image in an unknown gallery identity set (see Sec.~\ref{sec:setting}),
since it can be adopted by unauthorized applications for recognizing the identity information of users. As stated in Fig.~\ref{fig:intro}, \emph{face encryption} is to block the ability of automatic inference on malicious applications, making them predict a wrong authorized or virtual target by the service providers. 
In general, little is known about the face recognition system and no direct query access is possible. 
Therefore, we need to generate adversarial masks against a surrogate known model with the purpose of deceiving a \textit{black-box} face recognition system. Moreover, we try to not affect the user experience when users share the protected photos on social media, and simultaneously conceal their identities from unauthorized recognition systems. Thus, the protected images should also be visually natural from the corresponding original ones, otherwise it may introduce undesirable artifacts as a result.



To address the aforementioned challenges, we propose a \textbf{targeted identity-protection iterative method} (TIP-IM) for face encryption against black-box face recognition systems. The proposed method generates adversarial identity masks that are both 
transferable and imperceptible. A good transferability implies that a model can 
effectively deceive other black-box face recognition systems, meanwhile the imperceptibility means that a photo manipulated by an adversarial identity mask is visually natural for the human observers. 
Specifically, to ensure the generated images are not arbitrarily misclassified as other identities, we randomly choose a set of face images from a dataset collected from the internet as the specified targets in our experiments\footnote{We choose face images from the Internet only for experimental illustration, which simulates the authorized or generated target set of face images provided by the service providers.}. 
Our method obtains superior performance against white-box and black-box face systems with multiple target identities 
via a novel iterative optimization algorithm.

Extensive experiments under practical and challenging open-set~\footnote{The testing identities are disjoint from the
training sets, which is regarded as the opposite of close-set classification task.} test scenarios~\cite{liu2017sphereface} demonstrate that our algorithm provides 95+\% protection success rate against white-box face systems, and outperforms previous methods by a margin even against various state-of-the-art algorithms. Besides, we also demonstrate its effectiveness in a real-world experiment by considering a commercial API service.
Our main contributions are summarized as
\begin{itemize}
    \item We involve some valuable considerations to protect privacy against unauthorized identification systems from the user's perspective, including targeted protection, natural outputs, black-box face systems, and unknown gallery set.
    \item We propose a targeted identity-protection iterative method (TIP-IM) to generate an adversarial identity mask, in which we consider multi-target sets and introduce a novel optimization mechanism to guarantee effectiveness under various scenarios.


\end{itemize}


\begin{table*}[t]
\scriptsize 
\setlength{\tabcolsep}{5pt}
\begin{center}
\begin{tabular}{c|ccccccccc}
\hline
& \tabincell{c}{Evasion~\cite{wilber2016can}} & \tabincell{c}{PP-GAN~\cite{wu2018privacy}} & \tabincell{c}{Inpainting~\cite{sun2018natural}} & \tabincell{c}{Replacement~\cite{sun2018hybrid}} & \tabincell{c}{Eyeglasses~\cite{Sharif2016Accessorize}} & \tabincell{c}{Evolutionary~\cite{dong2019efficient}} & \tabincell{c}{LOTS~\cite{rozsa2017lots}} & \tabincell{c}{GAMAN~\cite{oh2017adversarial}} & Ours \\
\hline
\tabincell{c}{Unknown gallery set} & No & No & No & No & No & \textbf{Yes} & No & No & \textbf{Yes}\\

\tabincell{c}{Target identity} & No & No & No & No & \textbf{Yes} & \textbf{Yes} & \textbf{Yes} & \textbf{Yes} & \textbf{Yes}\\

\tabincell{c}{Black-box model} & \textbf{Yes} & \textbf{Yes} & \textbf{Yes} & \textbf{Yes} & No & \emph{Yes (Queries)} & No & No & \textbf{Yes}\\

\tabincell{c}{Natural outputs} & No & \textbf{Yes} & \textbf{Yes} & \textbf{Yes} & No & No & \textbf{Yes} & No & \textbf{Yes}\\

\tabincell{c}{Same faces} & \emph{Partially} & \emph{Partially} & No & No & \textbf{Yes} & \textbf{Yes} & \textbf{Yes} & \textbf{Yes} & \textbf{Yes}\\
\hline
\end{tabular}
\end{center}
\vspace{-2ex}
\caption{A comparison among different methods w.r.t the unknown gallery set, targeted misclassification of the output faces, black-box face models, natural outputs, and whether the output faces are recognized as the same identities as the original ones for human observers. }
\label{tab:related-work}
\vspace{-3ex}
\end{table*}

\section{Related Work}
In this section, we review related work on face encryption. Typical information encryption~\cite{lu2009face,singh2013study,li2010secure} requires to encode message to protect information from exposed by unauthorized parties, whereas face encryption aims to protect users' facial information from being unconsciously identified and exposed by the  unauthorized AI recognition systems. We provide a comprehensive comparison between the previous methods and ours in Tab.~\ref{tab:related-work}. 

\textbf{Obfuscation-based methods.} Several works have been developed to protect private identity information in personal photos against face or person recognition systems.
Earlier works~\cite{wilber2016can,ribaric2016identification} study the performance of these systems under various simple image obfuscation methods, such as blurring, pixelation, darkening, occlusion, \emph{etc}. These methods have been shown to be ineffective against the current recognition systems~\cite{oh2016faceless,mcpherson2016defeating,li2019anonymousnet}, since they can adapt to the obfuscation patterns. More sophisticated techniques have been proposed thereafter. For example, generative adversarial networks (GANs)~\cite{goodfellow2014generative} provide a useful way to synthesize realistic images on the data distribution for image obfuscation~\cite{wu2018privacy}.
In~\cite{sun2018natural}, the obfuscated images are generated by head in-painting conditioned on the detected face landmarks.  
However, these image obfuscation methods often change the visual appearances of face images and even lead to unnatural outputs, limiting their utility for users. 

\textbf{Adversarial methods.}
Deep neural networks are susceptible to adversarial examples~\cite{szegedy2013intriguing,goodfellow2014explaining,dong2019benchmarking,pang2020boosting}, so are the face recognition models~\cite{Sharif2016Accessorize,dong2019efficient,yang2020design}.  
Fawkes~\cite{shan2020fawkes} fools unauthorized facial recognition models by introducing adversarial examples into training data. A recent work~\cite{oh2017adversarial} proposes to craft protected images from a game theory perspective. However, our work is different from their previous works in three aspects. First, 
we focus on the unknown face systems without changing training data~\cite{shan2020fawkes}, while~\cite{oh2017adversarial} assumes the white-box access to the target model. Second, we consider the open-set face identification protocol with an unknown gallery set rather than a closed-set classification scenario. Ours can provide better  protection  success  rate  against unknown recognition systems on more practical open-set  scenarios.
Third, we have the ability to control the naturalness of the protected images under the $\ell_{p}$ norm.

\textbf{Differential privacy.}
As one of the popular definitions of privacy, differential privacy (DP)~\cite{dwork2008differential,dwork2006calibrating} has been introduced in the context of machine learning and data statistics, which requires that the returned information about an underlying dataset is robust to any change of one individual, thus protecting the privacy of entities. Along this routine, many promising DP techniques~\cite{dwork2006calibrating,mcsherry2007mechanism} and practical applications~\cite{blum2005practical,barak2007privacy} on DP have been developed. While DP withholds the existence of entities in a dataset, in this paper we focus on concealing the identity of a single one by exploiting the vulnerability of neural networks~\cite{szegedy2013intriguing}.

\section{Adversarial Identity Mask}
\label{sec:setting}

Let $f(\bm{x}): \mathcal{X}\rightarrow\mathbb{R}^d$ denote a face recognition model that extracts a fixed length feature representation in $\mathbb{R}^d$ for an input face image $\bm{x}\in\mathcal{X}\subset\mathbb{R}^n$.
Given the metric $\mathcal{D}_f(\bm{x}_1,\bm{x}_2) = \|f(\bm{x}_1) - f(\bm{x}_2)\|_2^2$ that measures the feature distance between two face images, face recognition compares the distance between a probe image and a gallery set of face images $\mathcal{G}=\{\bm{x}_1^{g}, ..., \bm{x}_m^g\}$, and returns the identity whose face image has the nearest feature distance with the probe image.

In this paper, 
we involve some valuable considerations from the user's perspective, to protect user's photos against an illegal face recognition systems, as illustrated in Fig.~\ref{fig:intro}. 
Specifically,
to conceal the true identity $y$ of a user's image $\bm{x}^{r}$, we aim to generate a protected image $\bm{x}^{p}$ by adding an adversarial identity mask $\bm{m}^{a}$ to $\bm{x}^{r}$ 
which can be denoted by $\bm{x}^{p}=\bm{x}^{r}+\bm{m}^{a}$ to make the face recognition system predict $\bm{x}^{p}$ as a different authorized identity or virtual identity corresponding to a generated image. 
Rather than specifying a single target identity for generating the protected image, we choose an identity set $\mathcal{I}=\{y_1, ..., y_k\}$, i.e., we allow the face recognition system to recognize the protected image as an arbitrary one of the target identities in $\mathcal{I}$ rather than a single one, which makes identity protection easier to achieve due to the relaxed constraints.

Formally, let $\mathcal{G}_y=\{\bm{x}|\bm{x}\in\mathcal{G}, \mathcal{O}(\bm{x})=y\}$ denote a subset of $\mathcal{G}$ containing all face images belonging to the true identity $y$ of $\bm{x}^{r}$, with $\mathcal{O}$ being an oracle to give the ground-truth identity labels, and 
$\mathcal{G}_{\mathcal{I}}=\bigcup\limits_{1\leq i\leq k}\mathcal{G}_{y_i}$ denote the face images belonging to the target identities of $\mathcal{I}$ in the gallery set $\mathcal{G}$. 
To conceal the identity of $\bm{x}^{r}$, the protected image $\bm{x}^{p}$ should satisfy the constraint as
\begin{equation}
\label{eq:goal}
    \exists \bm{x}^t\in \mathcal{G}_{\mathcal{I}}, \forall \bm{x} \in \mathcal{G}_y: \mathcal{D}_f(\bm{x}^{p},\bm{x}) > \mathcal{D}_f(\bm{x}^{p},\bm{x}^t). \\
\end{equation}
It ensures that the feature distance between the generated protected image $\bm{x}^{p}$ and a target identity's image $\bm{x}^t$ in $\mathcal{G}_{\mathcal{I}}$ is smaller than that between $\bm{x}^{p}$ and any image belonging to the true identity $y$ in $\mathcal{G}_y$.



We involve more practical considerations from a general user's perspective than the previous studied setting, in the following three aspects. 

\begin{figure}[t]
\centering
\includegraphics[width=0.8\linewidth]{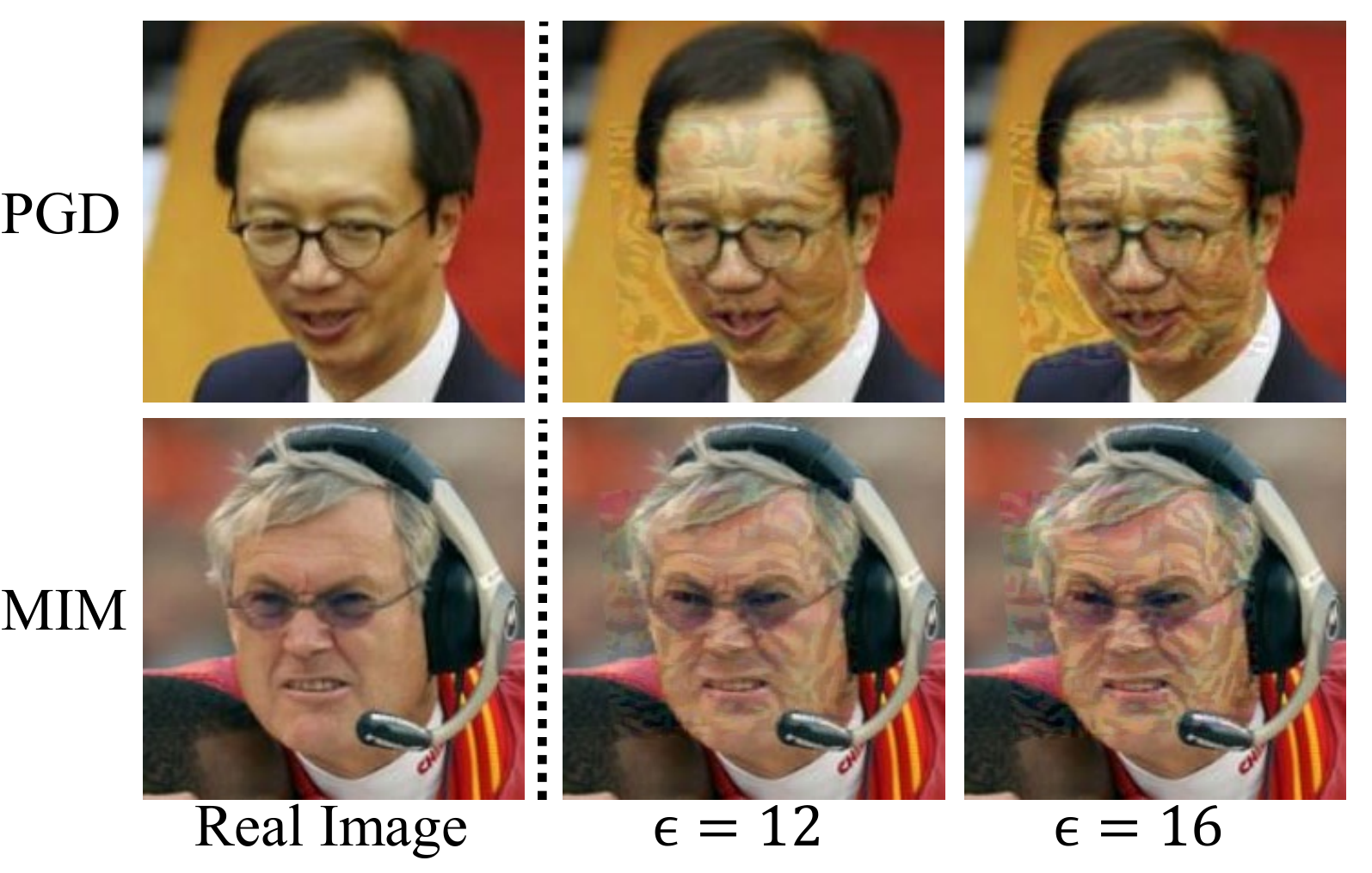}
\vspace{-1ex}
\caption{Illustration of different perturbations under the $l_\infty$ norm. More examples are presented in Appendix {\color{red} D}.}
\label{fig:perturbation}
\vspace{-2ex}
\end{figure}

\textbf{Naturalness.}  To make the protected image indistinguishable from the corresponding original one, a common practice is to restrict the $\ell_p$ ($p=2,\infty$, \emph{etc}.) norm between the protected and original examples, as $\|\bm{m}^{a}\|_p\leq\epsilon$.
However, the perturbation under the $\ell_p$ norm can not naturally fit human perception well~\cite{zhao2017generating,sen2019should}, as also illustrated in Fig.~\ref{fig:perturbation}. Therefore, we require that the protected image should look natural besides the constraint of the $\ell_p$ norm bound, to make it constrained on the data manifold of real images~\cite{stutz2019disentangling}, thus achieving imperceptible for human eyes. We use an objective function to promote the naturalness of the protected image, which will be specified in the following section.

\textbf{Unawareness of gallery set.} 
For a real-world face recognition system, we have no knowledge of its gallery set $\mathcal{G}$, meaning that we are not able to solve Eq.~\eqref{eq:goal} directly, while previous works assume the availability of the gallery set or a closed-set protocol (i.e., no gallery set). To address this issue, we use substitute face images for optimization. In particular, we collect an image set $\hat{\mathcal{G}}_{\mathcal{I}}$ containing face images that belong to the target identities of $\mathcal{I}$ as a surrogate for $\mathcal{G}_{\mathcal{I}}$; and use $\{\bm{x}^{r}\}$ directly instead of $\mathcal{G}_y$. The rationality of using substitute images is that face representations of one identity are similar, and thus the representation of a protected image optimized to be similar to the substitutes can also be close to images belonging to the same target identity in the gallery set.

\textbf{Unknown face systems.} 
In practice, we are also unaware of the face recognition model, include its architecture, parameters, and gradients. Previous methods rely on  the white-box access to the target model, which are impractical in real-world scenarios for identity protection. Thus we adopt a surrogate white-box model against which the protected images are generated, with the purpose of improving the transferability of adversarial masks against unknown face systems.

In summary, our considerations are designed to simulate the real-world scenarios with minimum assumptions of the target face recognition system, which is also more challenging than previously studied settings.

\section{Methodology}
\label{sec:method}

To achieve the above requirements, we propose a \textbf{targeted identity-protection iterative method} (TIP-IM) to generate protected images in this section. 

\subsection{Problem Formulation}
\vspace{-1ex}
\label{sec:problem}
To generate a protected image $\bm{x}^{p}$ that is both effective for obfuscation against face recognition systems and visually natural for human eyes, we formalize the objective of targeted privacy-protection function as
\begin{equation}
\small
\label{eq:objective}
\begin{gathered}
\min_{ \bm{x}^t,\bm{x}^{p}} \mathcal{L}_{iden}(\bm{x}^t, \bm{x}^{p}) =   \mathcal{D}_f(\bm{x}^{p}, \bm{x}^t) - \mathcal{D}_f(\bm{x}^{p}, \bm{x}^{r}) \\
\text{s.t.}\; \|\bm{x}^{p}-\bm{x}^{r}\|_p\leq\epsilon,\mathcal{L}_{nat}(\bm{x}^{p})\leq\eta,
\end{gathered}
\end{equation}
where $\bm{x}^t\in\hat{\mathcal{G}}_{\mathcal{I}}$, and $\mathcal{L}_{iden}$ is a relative identification loss that enables the generated $\bm{x}^{p}$ to increase the distance gap between a targeted image $\bm{x}^t$ and the original image $\bm{x}^{r}$ in the feature space. 
$\mathcal{L}_{nat}\leq \eta$ is a constraint condition that  makes $\bm{x}^{p}$ look natural. We also restrict the $\ell_p$ norm of the perturbation to be smaller than a constant $\epsilon$ such that the visual appearance does not change significantly.
Note that for the unawareness of gallery set, we use the substitute face images $\hat{\mathcal{G}}_{\mathcal{I}}$ in our objective~\eqref{eq:objective}; for the unknown model, we generate a protected image $\bm{x}^{p}$ against a surrogate white-box model with the purpose of fooling the black-box model based on the transferability. Thus the requirements in the proposed targeted identity-protection function can be fulfilled by solving Eq.~\eqref{eq:objective}.


Although the perturbation is somewhat small due to the $\ell_p$ norm constraint in Eq.~\eqref{eq:objective}, it can be still perceptible and not natural for human eyes, as shown in Fig.~\ref{fig:perturbation} . Therefore, we add a loss $\mathcal{L}_{nat}$ into our objective to explicitly encourage the naturalness of the generated protected image.
In this paper, we adopt the \emph{maximum mean discrepancy} (MMD)~\cite{borgwardt2006integrating} as $\mathcal{L}_{nat}$, because it is an effective non-parametric and differentiable metric capable of comparing two data distributions and evaluating the imperceptibility of the generated images.
In our case, given two sets of data $\mathbf{X}^{p}=\{\bm{x}^{p}_1,...,\bm{x}^{p}_N\}$ and $\mathbf{X}^{r}=\{\bm{x}^{r}_1,...,\bm{x}^{r}_N\}$ comprised of $N$ generated data and $N$ real images,
MMD calculates their discrepancy by
\begin{equation}
\small
\label{eq:mmd_em}
    \mathrm{MMD}(\mathbf{X}^{p},\mathbf{X}^{r}) =  \big|\big|\frac{1}{N} \sum_{i=1}^{N}\phi(\bm{x}_{i}^{p}) - \frac{1}{N} \sum_{j=1}^{N}\phi(\bm{x}_{j}^{r})\big|\big|_{\mathcal{H}}^{2},
\end{equation}
where $\phi(\cdot)$ maps the data to a reproducing kernel Hilbert space (RKHS)~\cite{borgwardt2006integrating}. We adopt the same $\phi(\cdot)$ as in~\cite{borgwardt2006integrating}. By minimizing MMD between the samples $\mathbf{X}^{p}$ from the generated distribution and the samples $\mathbf{X}^{r}$ from the real data distribution, we can constrain $\mathbf{X}^{p}$ to lie on the manifold of real data distribution, meaning the protected images in $\mathbf{X}^{p}$ will be as natural as real examples.

Since MMD is a differentiable metric and defined on the batches of images, we thus integrate MMD into Eq.~\eqref{eq:objective} and rewrite our objective with a batch-based formulation\footnote{In case of there is only one single image or a small number of images, we can augment the images with multiple transformations to make up a large batch, with the results shown in Appendix {\color{red} B}.} as 
\begin{equation}
\label{eq:total}
\small
\begin{gathered}
 \min_{\textbf{X}^{p}}  \mathcal{L}(\mathbf{X}^{p}) = \frac{1}{N}\sum_{i=1}^{N} \mathcal{L}_{iden}(\bm{x}^t_i, \bm{x}^{p}_i)  + \gamma \cdot \mathrm{MMD}(\mathbf{X}^{p}, \mathbf{X}^{r}), \\ \text{s.t.}\;   \|\bm{x}^{p}_i-\bm{x}^{r}_i\|_p\leq\epsilon,
 \end{gathered}
\end{equation}
where $\bm{x}^t_i\in\hat{\mathcal{G}}_{\mathcal{I}}$ and $\gamma$ is a hyperparameter to balance these two losses.



    
    
    

\subsection{Targeted Identity-Protection Iterative Method}

Given the overall loss $\mathcal{L}(\mathbf{X}^{p})$ in Eq.~\eqref{eq:total}, we can therefore generate the batch of protected images $\mathbf{X}^{p}$ by minimizing $\mathcal{L}(\mathbf{X}^{p})$. Given the definitions of $\mathcal{L}_{iden}$ and MMD in Eq.~\eqref{eq:objective} and Eq.~\eqref{eq:mmd_em}, $\mathcal{L}(\mathbf{X}^{p})$ is a differentiable function w.r.t. $\mathbf{X}^{p}$, and thus we can iteratively apply fast gradient method~\cite{Kurakin2016} multiple times with a small step size $\alpha$ to generate protected images by minimizing the loss $\mathcal{L}(\mathbf{X}^{p})$.
In particular, we optimize $\mathbf{X}^{p}$ via
\begin{equation}
 \label{eq:update}
 \small
     \mathbf{X}^{p}_{t+1} = \Pi_{\{\mathbf{X}^{r},\ell_p, \epsilon\}}\big( \mathbf{X}_{t}^{p} - \alpha \cdot \mathrm{Normalize}(\nabla_{\mathbf{X}}\mathcal{L}(\mathbf{X}_{t}^{p}))  \big),
 \end{equation}
where $\mathbf{X}^{p}_{t}$ is the batch of protected images at the $t$-th iteration, $\Pi$ is the projection function that projects the protected images onto the $\ell_p$ norm bound, and $\mathrm{Normalize}(\cdot)$ is used to normalize the gradient (e.g., a sign function under the $\ell_{\infty}$ norm bound or the $\ell_2$ normalization under the $\ell_2$ norm bound). We perform the iterative process for a total number of $T$ iterations and get the final protected images as $\mathbf{X}^{p}_T$.
To prevent protected images from falling into local minima and improve their transferability for other black-box face recognition models, we incorporate the momentum technique~\cite{Dong2017} into the iterative process.

\subsection{Search Optimal $\bm{x}^t$ via Greedy Insertion}
\label{sec:multi-target}
When there is only one target face image in $\hat{\mathcal{G}}_{\mathcal{I}}$, we do not need to consider how to select $\bm{x}^t$ to effectively incorporate into Eq.~\eqref{eq:total}. 
When the target set $\hat{\mathcal{G}}_{\mathcal{I}}$ contains multiple target images, it offers more potential optimization directions to get better performance. Therefore, we develop an optimization algorithm to search for the optimal target while generating protected images as Eq.~\eqref{eq:update}.
Specifically, for the iterative procedure in Eq.~\eqref{eq:update} with $T$ iterations, we select a representative target for each protected image in $\hat{\mathcal{G}}_{\mathcal{I}}$ at each iteration for updates, which belongs to a subset selection problem.


\textbf{Definition 1.} \textit{Let $S_t$ denote the set of the selected targets from $\hat{\mathcal{G}}_{\mathcal{I}}$ at each iteration until the $t$-th iteration. Let $F$ denote a set mapping function that outputs a gain value (larger is better) in $\mathbb{R}$ for a set. For $\bm{x}^t \in \hat{\mathcal{G}}_{\mathcal{I}}$, we define $\Delta(\bm{x}^t | S_t) = F(S_t \cup \{\bm{x}^t\}) - F(S_t)$ be the marginal gain of $F$ at $S_t$ given $\bm{x}^t$}. 


%

\begin{algorithm}[t]
    \caption{Search Optim. via Greedy Insertion}\label{algo2}
\KwInput{The privacy-protection objective function $\mathcal{L}_{iden}$ from Eq.~(\ref{eq:objective}); a real face $\bm{x}^{r}$ and a multi-identity face images $\hat{\mathcal{G}}_{\mathcal{I}}$; a feature representation function $f$; a gain function $\mathrm{G}$.}
\KwInput{The protected image $\bm{x}^{p}$ generated before the current iteration.}
\KwOutput{The best target image ${\bm{x}^t}^*$ in $\hat{\mathcal{G}}_{\mathcal{I}}$.}
    {
    $g_{best}\leftarrow 0$; ${\bm{x}^t}^*\leftarrow \mathrm{None}$;\\
    \For{$\bm{x^t}$ in $\hat{\mathcal{G}}_{\mathcal{I}}$}
        {
        Get the loss $\mathcal{L}_{iden}(\bm{x^t}, \bm{x}^{p})$ via Eq.~\eqref{eq:objective}; \\
        Compute the gradient $ \nabla_{\bm{x}} \mathcal{L}_{iden}(\bm{x^t}, \bm{x}^{p})$; \\
        Generate candidate protected image $\hat{\bm{x}}^{p} = \Pi_{\{\bm{x}^{r}, \ell_p, \epsilon\}}(\bm{x}^{p} - \alpha\cdot \mathrm{Normalize}(\nabla_{\bm{x}} \mathcal{L}_{iden}(\bm{x^t}, \bm{x}^{p})))$;\\
        Calculate $g = \mathrm{G}(\hat{\bm{x}}^{p})$;\\
        \If{$g > g_{best}$}
        {
        $g_{best} \leftarrow g$;
        ${\bm{x}^t}^* \leftarrow \bm{x}^t$;
        }
        }
    
    }
\end{algorithm}

Formally, as the iteration gets increasing in the iteration loop, if the marginal gain decreases monotonically, then $F$ will belong to the family of submodular functions~\cite{zhou2016causal}. For a submodular problem, a greedy algorithm can be used to find an approximate solution, and it has been shown that submodularity will have a ($1-1/e$)-approximation~\cite{nemhauser1978analysis} for monotinely submodular functions. Although our iterative identity-protection method is not guaranteed to be strictly submodular, the solution based on greedy insertion still plays an obvious role even if submodularity is not strictly decreased~\cite{zhou2016causal,feige2011maximizing},  which evaluate the theoretical results and justify that the greedy algorithm has a performance guarantee for the maximization of approximate submodularity. Therefore, we adopt the greedy insertion solution as an approximately optimal solution for our multi-target problem. 

By analyzing the above setting, we perform the approximate submodular optimization by greedy insertion algorithm, which calculates the gain of every object from the target set at each iteration and integrates the object with the largest gain into current subset $S_t$ by Definition $1$ as  
\begin{equation}
\label{blacktars}
\small
     S_{t+1} = S_{t} \cup \{\argmax_{\bm{x}^t\in \hat{\mathcal{G}}_{\mathcal{I}}} \Delta (\bm{x}^t|S_{t})\}.
\end{equation}
To achieve this, we need to define the above set mapping function $F$. In particular, we specify $F$ as first generating a protected image $\bm{x}^{p}$ given the targets in $S_t$ for $t$ iterations via Eq.~\eqref{eq:update}, and then using a function $\mathrm{G}$ to compute a gain value by Definition $1$. An appropriate gain function $\mathrm{G}$ should choose examples that are effective for minimizing $\mathcal{L}_{iden}(\bm{x}^t,\bm{x}^{p})$ at each iteration. It is noted that $\mathrm{G}$ must also have positive value and a larger value indicates better performance. Based on this analysis, we design a feature-based similarity gain function as
\begin{equation}
\label{1eq:gain3}
\small
     \mathrm{G}(\bm{x}^{p}) = \log \big( 1 + \mathop{\max}\limits_{\bm{x^t} \in \hat{\mathcal{G}}_{\mathcal{I}}} \exp(\mathcal{D}_f(\bm{x}^{p}, \bm{x}^{r})- \mathcal{D}_f(\bm{x}^{p}, \bm{x}^t))\big),
\end{equation}
where the algorithm tends to select a target closer to the real image in the feature space at each iteration. The algorithm is summarized in Algorithm~\ref{algo2}.

\section{Experiments}
In this section, we conduct extensive experiments in the aspect of identity protection to demonstrate the effectiveness of the proposed method. We thoroughly evaluate different properties of our method based on various state-of-the-art face recognition models.\footnote{Code at \url{https://github.com/ShawnXYang/TIP-IM}.}

\vspace{-0.1cm}
\subsection{Experimental Settings}

\label{sec:exps_setting}
\textbf{Datasets.} The experiments are constructed on the Labeled Face in the Wild  (LFW)~\cite{huang2008labeled} and MegFace~\cite{kemelmacher2016megaface} datasets. We involve some additional considerations to draw near realistic testing scenarios:
1) \textbf{practical gallery set}: 
    we first select $500$ different identities as the protected identities. Meanwhile, we randomly select an image from each identity as the probe image (total $500$ images), and the \emph{other} images (not selecting one template) for each identity are assembled to form a gallery set because the gallery of face encryption in social media includes multiple images per identity (more difficult yet practical for simultaneously concealing multiple images per identity);
2) \textbf{target identities}: we randomly select another $10$ identities as $\mathcal{I}$ from a dataset in the Internet named MS-Celeb-1M~\cite{guo2016ms}. We select one image for each of these target identities to form $\hat{\mathcal{G}}_{\mathcal{I}}$ and the remaining images are integrated into the gallery set, which can ensure the unawareness of gallery set that target images in the optimized phase are different from ones in the testing;
3) \textbf{additional identities:}  we add additional $500$ identities to the gallery set, which accords with a realistic test scenario.  
Thus we construct two challenging yet practical data scenarios (over 1k identities and total 10K images). 

\textbf{Target models.} We select models with diverse backbones and training losses to fully demonstrate the ability to protect user privacy in Tab.~\ref{tab:model}. In experiments, we first use MTCNN~\cite{zhang2016joint} to detect faces in the image, then align the images and crop them to $112\times112$, meaning that the identity masks are executed only in the face area. Only one model is used as known model to generate the identity masks, and test the protection performance in other unknown models.

\begin{table}[t]
    \begin{center}
    \scriptsize 
    \setlength{\tabcolsep}{6pt}
    \begin{tabular}{c|ccc}
    \hline
        Model & Backbone & Loss & Parameters (M)\\
         \hline
        FaceNet~\cite{schroff2015facenet} & InceptionResNetV1 & Triplet & 27.91\\
	 SphereFace~\cite{liu2017sphereface} & Sphere20 & A-Softmax &28.08 \\ 
	 CosFace~\cite{wang2018cosface} & Sphere20 & LMCL & 22.67 \\
	    ArcFace~\cite{deng2019arcface} & IR-SE50 & Arcface & 43.80\\
		 MobleFace~\cite{chen2018mobilefacenets} & MobileFaceNet & Softmax & 1.20\\
		 ResNet50~\cite{he2016} & ResNet50 & Softmax & 40.29 \\

         \hline
    \end{tabular}
    \end{center}
    \vspace{-1ex}
    \caption{Chosen target models that lie in various settings, including different architectures and training objectives.}
    \label{tab:model}
    \vspace{-2ex}
\end{table}

\begin{table*}[t]
    \begin{center}
    \scriptsize 
    \setlength{\tabcolsep}{8pt}
    \begin{tabular}{c|c|cc|cc|cc|cc|cc|cc}
    \hline
        \multirow{2}{*}{} & \multirow{2}{*}{Method} & 
        \multicolumn{2}{c|}{ArcFace} & \multicolumn{2}{c|}{MobileFace} & \multicolumn{2}{c|}{ResNet50} & \multicolumn{2}{c|}{SphereFace} &
        \multicolumn{2}{c|}{FaceNet} & \multicolumn{2}{c}{CosFace}\\
         \cline{3-14}
         & & R1-T & R5-T & R1-T & R5-T & R1-T & R5-T & R1-T & R5-T & R1-T & R5-T & R1-T & R5-T \\
         \cline{1-14}
         {\rotatebox[origin=c]{0}{ArcFace}}& \tabincell{c}{MIM~\cite{Dong2017} \\ DIM~\cite{xie2019improving} \\ MT-DIM~\cite{xie2019improving} \\ Center-Opt \\ TIP-IM} &  
         \tabincell{c}{94.0$^*$ \\94.8$^*$\\34.8$^*$\\59.4$^*$\\\textbf{97.2$^*$}} &
         \tabincell{c}{96.9$^*$ \\97.6$^*$\\68.2$^*$\\84.6$^*$\\\textbf{98.8$^*$}} &
         \tabincell{c}{14.3 \\16.8\\18.8\\36.8\\\textbf{69.8}} &
         \tabincell{c}{45.8 \\48.0\\53.6\\66.0\\\textbf{90.6}} &
         \tabincell{c}{8.2\\10.8\\15.8\\28.8\\\textbf{56.0}} &
         \tabincell{c}{32.4\\34.8\\46.0\\57.6\\\textbf{80.6}} & 
         \tabincell{c}{3.1\\4.2\\3.8\\6.6\\\textbf{13.2}} &
         \tabincell{c}{14.5\\15.6\\18.4\\21.4\\\textbf{32.0}} &
         \tabincell{c}{3.1\\4.4\\9.6\\11.8\\\textbf{32.8}} &
         \tabincell{c}{17.9\\19.0\\33.6\\35.4\\\textbf{56.2}} & \tabincell{c}{1.7\\2.6\\2.0\\3.8\\\textbf{11.4}} & 
         \tabincell{c}{10.1\\11.0\\11.2\\13.0\\\textbf{31.0}}\\
         \hline
         {\rotatebox[origin=c]{0}{MobileFace}}& \tabincell{c}{MIM~\cite{Dong2017} \\DIM~\cite{xie2019improving} \\ MT-DIM~\cite{xie2019improving} \\ Center-Opt \\ TIP-IM} & 
         \tabincell{c}{8.1\\9.4\\10.6\\14.8\\\textbf{44.0}} &
         \tabincell{c}{27.9\\28.8\\30.2\\41.8\\\textbf{68.2}} &
         \tabincell{c}{96.1$^*$ \\\textbf{96.6$^*$}\\40.2$^*$\\53.0$^*$\\\textbf{96.6$^*$}} &
         \tabincell{c}{98.3$^*$ \\98.4$^*$\\73.2$^*$\\83.4$^*$\\\textbf{99.2$^*$}} &
         \tabincell{c}{26.7\\28.8\\18.6\\21.8\\\textbf{62.8}} &
         \tabincell{c}{61.5\\63.2\\49.4\\53.6\\\textbf{85.8}} & \tabincell{c}{4.5\\6.2\\6.8\\5.8\\\textbf{12.6}} &
         \tabincell{c}{18.0\\19.2\\22.2\\25.6\\\textbf{29.8}} &
         \tabincell{c}{3.7\\4.8\\9.8\\12.2\\\textbf{28.8}} &
         \tabincell{c}{19.4\\20.6\\27.0\\29.6\\\textbf{46.2}} & \tabincell{c}{0.3\\1.2\\2.0\\3.6\\\textbf{12.4}} & \tabincell{c}{4.1\\5.2\\11.0\\12.4\\\textbf{31.2}} \\
         \hline
         {\rotatebox[origin=c]{0}{ResNet50}}& \tabincell{c}{MIM~\cite{Dong2017} \\DIM~\cite{xie2019improving} \\ MT-DIM~\cite{xie2019improving} \\ Center-Opt \\ TIP-IM} & 
         \tabincell{c}{5.2\\7.0\\14.2\\13.2\\\textbf{34.2}} &
         \tabincell{c}{24.7\\26.4\\37.6\\37.14\\\textbf{56.8}} &
         \tabincell{c}{24.6\\26.6\\22.4\\26.8\\\textbf{62.4}} &
         \tabincell{c}{56.5\\57.2\\52.6\\58.6\\\textbf{83.4}} &
         \tabincell{c}{30.1$^*$\\31.4$^*$\\31.4$^*$\\41.6$^*$\\\textbf{95.6$^*$}} &
         \tabincell{c}{64.8$^*$\\65.0$^*$\\65.0$^*$\\73.4$^*$\\\textbf{98.2$^*$}} &
         \tabincell{c}{8.1\\9.2\\5.2\\6.8\\\textbf{11.4}} &
         \tabincell{c}{23.4\\24.2\\16.8\\21.0\\\textbf{25.6}} &
         \tabincell{c}{4.9\\6.8\\9.4\\9.2\\\textbf{23.2}} &
         \tabincell{c}{20.7\\22.8\\29.2\\27.2\\\textbf{40.0}} &
         \tabincell{c}{0.9\\2.0\\1.8\\2.4\\\textbf{10.8}}  & \tabincell{c}{5.6\\6.6\\8.8\\12.0\\\textbf{26.2}}\\

         \hline
    \end{tabular}
    \end{center}
    \vspace{-1ex}
    \caption{Rank-1-T and Rank-5-T (\%) of black-box identity protection against different models on LFW. $^*$ indicates white-box results.}
    \label{tab:transfer}
    \vspace{-1ex}
\end{table*}

\textbf{Compared Methods.} 
We investigate many adversarial face encryption methods~\cite{Sharif2016Accessorize,oh2017adversarial}, which essentially depend on single-target adversarial attack method~\cite{Kurakin2016}. Advanced MIM~\cite{Dong2017} introduces the momentum into iterative process~\cite{Kurakin2016} to improve the black-box transferability, and DIM and TIM~\cite{xie2019improving,dong2019evading} aim to achieve better transferability by input or gradient diversity. Note that TIM only focus on evading defense models and experimentally also achieve worse performance than MIM and DIM. Thus MIM and DIM are regarded as more effective single-target black-box algorithms as comparison. As original DIM only support single-target attack in the iterative optimization, we thus incorporate a multi-target version for DIM via a dynamic assignment from same target set in the inner minimization, named MT-DIM.
Besides, we study the influence of other multi-target optimization methods. We denote an additional gain function based on Eq.~(\ref{1eq:gain3}) as $\mathrm{G}_{1}(\bm{x}) =\log \big( 1 + \sum_{\bm{x}^t \in \hat{\mathcal{G}}_{\mathcal{I}}}\exp(\mathcal{D}_f(\bm{x}, \bm{x}^{r} ) -\mathcal{D}_f(\bm{x},\bm{x}^t))\big)$ 
which is named \emph{Center-Opt}. Center-Opt promotes protected images to be updated towards the mean center of target identities in the feature space, which is similarly adopted in~\cite{rozsa2017lots}. Note that single-target methods calculate optimal result as final report by attempting a target from the same target set. We set the number of iterations as $N = 50$, the learning rate $\alpha = 1.5$ and the size of perturbation $\epsilon=12$ under the $\ell_{\infty}$ norm bound, which are identical for all the experiments.

\textbf{Evaluation Metrics.} 
To comprehensively evaluate the \textbf{protection success rate}, we report Rank-N targeted identity success rate named \emph{Rank-N-T} and untargeted identity success rate named \emph{Rank-N-UT} (higher is better), which are consistent with the evaluation of face recognition~\cite{deng2019arcface,wang2018cosface}. Specifically, given a probe image $\bm{x}$ and a gallery set $\mathcal{G}$ with at least one image of the same identity with $\bm{x}$, meanwhile $\mathcal{G}$ has images of target identities. The face recognition algorithm ranks the distance $\mathcal{D}_f$ for all images in the gallery to $\bm{x}$. Rank-N-T means that \emph{at least one} of the top N images belongs to the target identity, whereas Rank-N-UT needs to satisfy that top N images do not have the same identity as $\bm{x}$. In this paper, we report Rank-1-T / Rank-1-UT and Rank-5-T / Rank-5-UT. Note that Rank-1-T / Rank-1-UT (Accuracy / Misclassification) is the most common evaluation metric in prior works, whereas Rank-5-T / Rank-5-UT can provide a comprehensive understanding since it is not sure whether the image will reappear in the top-K candidates. All methods including single-target methods adopt the same target identities and evaluation criterion for a fair comparison.

To test the \textbf{imperceptibility} of the generated protected images, we adopt the standard quantitative measures---PSNR (dB) and structural similarity (SSIM)~\cite{wang2004image}, as well as MMD in the face area. For SSIM and PSNR, a larger value means better image quality, whereas a smaller MMD value indicates superior performance.




\begin{table*}[t]
    \begin{center}
    \scriptsize
    \setlength{\tabcolsep}{8pt}
    \begin{tabular}{c|c|cc|cc|cc|cc|cc|cc}
    \hline
        \multirow{2}{*}{} & \multirow{2}{*}{Attack} & 
        \multicolumn{2}{c|}{ArcFace} & \multicolumn{2}{c|}{MobileFace} &
        \multicolumn{2}{c|}{ResNet50} &
        \multicolumn{2}{c|}{SphereFace} & \multicolumn{2}{c|}{FaceNet} & \multicolumn{2}{c}{CosFace} \\
         \cline{3-14}
         & & R1-U & R5-U & R1-U & R5-U & R1-U & R5-U & R1-U & R5-U & R1-U & R5-U & R1-U & R5-U \\
         
         \hline
         
         {ArcFace}& \tabincell{c}{\tabincell{c}{DIM~\cite{xie2019improving} \\ MT-DIM~\cite{xie2019improving} \\ TIP-IM}} & 
         \tabincell{c}{95.8$^*$\\96.0$^*$\\ \textbf{97.4$^*$}} &
         \tabincell{c}{91.8$^*$\\93.6$^*$ \\\textbf{96.4$^*$}} &
         \tabincell{c}{67.6\\73.6\\\textbf{79.4}} &
         \tabincell{c}{58.2\\65.2\\\textbf{68.8}} &
         \tabincell{c}{58.2\\64.8\\\textbf{70.4}} &
         \tabincell{c}{48.0\\54.2\\\textbf{56.8}} &
         \tabincell{c}{79.6\\82.8\\\textbf{85.2}} &
         \tabincell{c}{68.2\\73.0\\\textbf{76.6}} &
         \tabincell{c}{67.4\\73.0\\\textbf{73.4}} &
         \tabincell{c}{53.6\\60.0\\\textbf{63.4}} &
         \tabincell{c}{74.2\\74.4\\\textbf{84.0}} &
         \tabincell{c}{62.8\\63.8\\\textbf{73.8}}
         \\
         \hline
         {MobileFace}& \tabincell{c}{DIM~\cite{xie2019improving} \\MT-DIM~\cite{xie2019improving} \\ TIP-IM} & 
         \tabincell{c}{60.2\\66.4\\\textbf{68.6}} &
         \tabincell{c}{45.4\\52.8\\\textbf{58.8}} &
         \tabincell{c}{96.4$^*$\\ 95.4$^*$\\\textbf{96.6$^*$}} &
         \tabincell{c}{92.0$^*$\\95.4$^*$\\\textbf{94.8$^*$}} &
         \tabincell{c}{72.2\\77.6\\\textbf{81.4}} &
         \tabincell{c}{59.0\\68.4\\\textbf{71.2}} &
          \tabincell{c}{80.0\\83.2\\\textbf{84.4}} &
         \tabincell{c}{69.6\\73.2\\\textbf{74.0}} &
         \tabincell{c}{68.4\\74.6\\\textbf{77.6}} &
         \tabincell{c}{53.6\\58.6\\\textbf{60.4}} &
         \tabincell{c}{75.6\\76.0\\\textbf{79.4}} &
         \tabincell{c}{62.2\\62.8\\\textbf{68.2}}\\
         \hline
         {ResNet50}& \tabincell{c}{DIM~\cite{xie2019improving} \\MT-DIM~\cite{xie2019improving} \\ TIP-IM} & \tabincell{c}{77.6\\79.0\\\textbf{83.6}} &
         \tabincell{c}{50.4\\52.4\\\textbf{59.6}} &
         \tabincell{c}{80.6\\84.8\\\textbf{87.0}} &
         \tabincell{c}{72.2\\75.2\\\textbf{81.8}} &
         \tabincell{c}{95.4$^*$\\94.4$^*$\\\textbf{96.8$^*$}} &
         \tabincell{c}{91.6$^*$\\93.8$^*$\\\textbf{94.6$^*$}} &
         \tabincell{c}{80.4\\82.2\\\textbf{85.8}} &
         \tabincell{c}{65.8\\72.4\\\textbf{75.0}} &
         \tabincell{c}{69.0\\77.8\\\textbf{79.4}} &
         \tabincell{c}{53.2\\60.8\\\textbf{65.4}} &
         \tabincell{c}{64.2\\77.0\\\textbf{83.6}} &
         \tabincell{c}{61.8\\65.8\\\textbf{73.0}}\\

         \hline
    \end{tabular}
    \end{center}
    \vspace{-1ex}
    \caption{Rank-1-UT and Rank-5-UT (\%) of black-box identity protection against different models on LFW.  $^*$ indicates white-box attacks.}
    \label{tab:transfer-u}
    \vspace{-2ex}
\end{table*}

\begin{table}[t]
    \begin{center}
    \scriptsize 
    \setlength{\tabcolsep}{5pt}
    \begin{tabular}{c|c|c|c|c|c}
    \hline
        & Metric & $\gamma=0.0$ & $\gamma=1.0$ & $\gamma=2.0$ & $\gamma=3.0$ \\
         \hline
         ArcFace& \tabincell{c}{PSNR($\uparrow$) \\ SSIM($\uparrow$) \\ MMD$(\downarrow$)} &  \tabincell{c}{25.26\\0.6520 \\0.7567} & 
         \tabincell{c}{25.59\\0.6690\\0.7562} & 
         \tabincell{c}{26.08\\0.6986\\0.7554} &
         \tabincell{c}{27.63\\0.7817\\0.7518} \\
         \hline
         MobileFace& \tabincell{c}{PSNR($\uparrow$) \\ SSIM($\uparrow$) \\ MMD($\downarrow$)} &  \tabincell{c}{25.24\\0.6490 \\0.7567} & 
         \tabincell{c}{25.18\\0.6523\\0.7568} & 
         \tabincell{c}{25.72\\0.6828\\0.7559} &
         \tabincell{c}{27.19\\0.7533\\0.7525} \\
         \hline
         ResNet50& \tabincell{c}{PSNR($\uparrow$) \\ SSIM($\uparrow$) \\ MMD($\downarrow$)} &  \tabincell{c}{25.14\\0.6507\\0.7570} & 
         \tabincell{c}{25.26\\0.6595\\0.7567} & 
         \tabincell{c}{25.74\\0.6897\\0.7558} &
         \tabincell{c}{27.21\\0.760\\0.7525} \\
         \hline
    \end{tabular}
    \end{center}
    \vspace{-1ex}
    \caption{The average PSNR (db), SSIM, and MMD of the protected images generated by TIP-IM with different $\gamma$.}
    \label{tab:nat}
    \vspace{-3ex}
\end{table}

\subsection{Effectiveness of Black-box Face Encryption}
\vspace{-0.1cm}

\begin{figure}[t]
\centering
\includegraphics[width=0.9\linewidth]{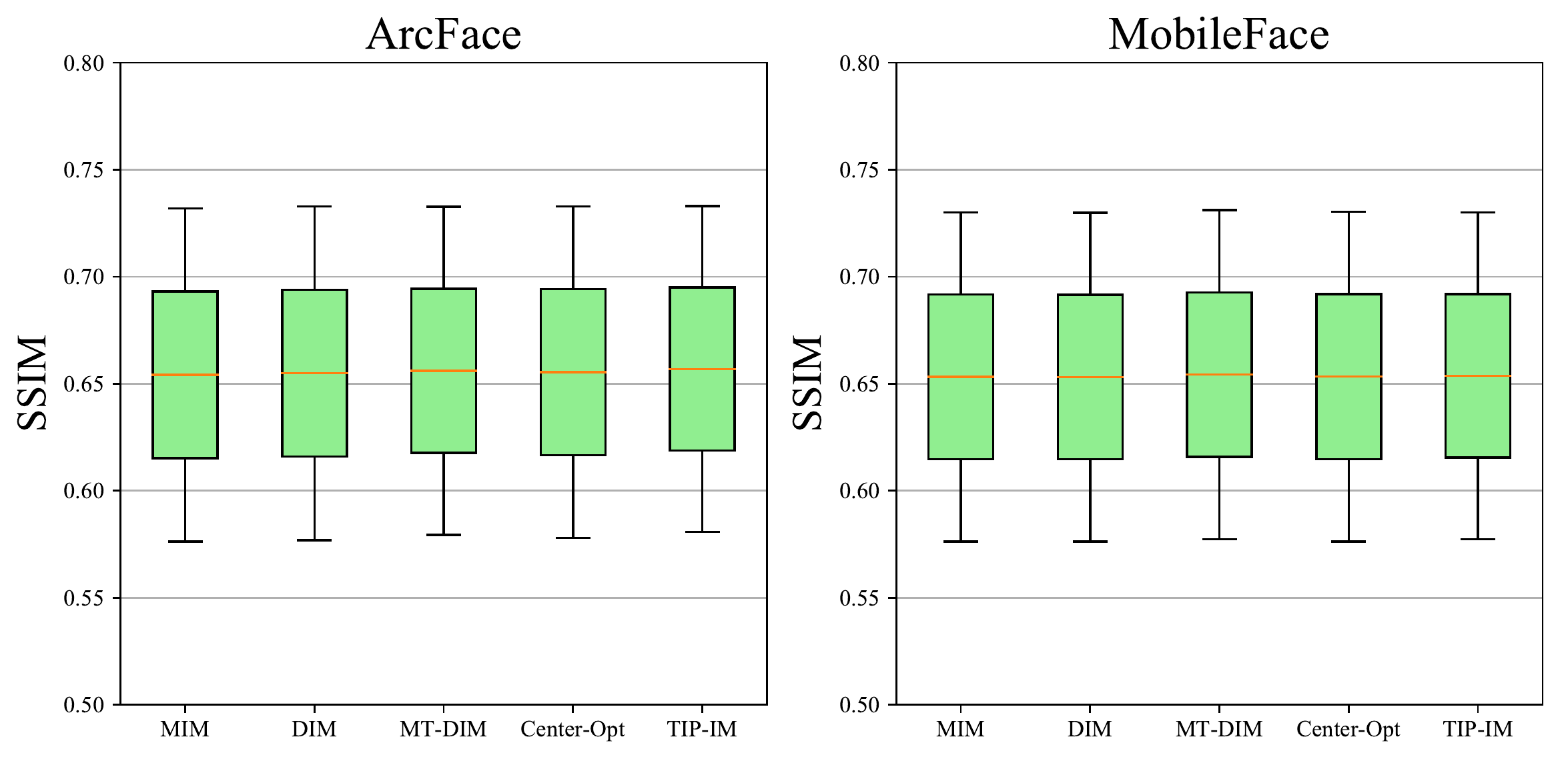}
\caption{Comparison of SSIM for different methods.}
\label{fig:ssim}
\vspace{-4ex}
\end{figure}

We first generate protected images against ArcFace, MobileFace, and ResNet50 respectively, by the proposed TIP-IM. We then feed the generated protected images to all face models for testing the performance in Tab.~\ref{tab:transfer} and Tab.~\ref{tab:transfer-u}. Our algorithm achieves nearly two times of the success rates than previous state-of-the-art method MT-DIM in terms of Rank-1-T and Rand-5-T, and outperforms other methods by a large margin, whereas SSIM values among compared methods are very similar in Fig.~\ref{fig:ssim}. MT-DIM obtains more acceptable performance than single method DIM, indicating that multi-target setting yields a better black-box transferability. It can be also observed that different multi-target methods will influence the performance, and proposed TIP-IM defined in Eq.~\eqref{1eq:gain3} achieves better performance than Center-Opt, . We also report the results of Rank-1-UT and Rank-5-UT in Tab.~\ref{tab:transfer-u}, which can still maintain the best performance with an average accuracy of Rank-1-UT over $\mathbf{80\%}$ for black-box models.  As a whole, TIP-IM provides more promising multi-target optimization direction, making generated protected images more effective for black-box models.  Note that the protected images generated 
by ArcFace have excellent transferability to the other black-box models.  Thus we will have priority to select ArcFace or ensemble mechanism~\cite{Dong2017} as the substitute model for better performance in practical application. Due to the space limitation, we leave the results of MegFace in Appendix {\color{red} A}.

         


\begin{figure*}[t]
\centering
\includegraphics[width=0.95\linewidth]{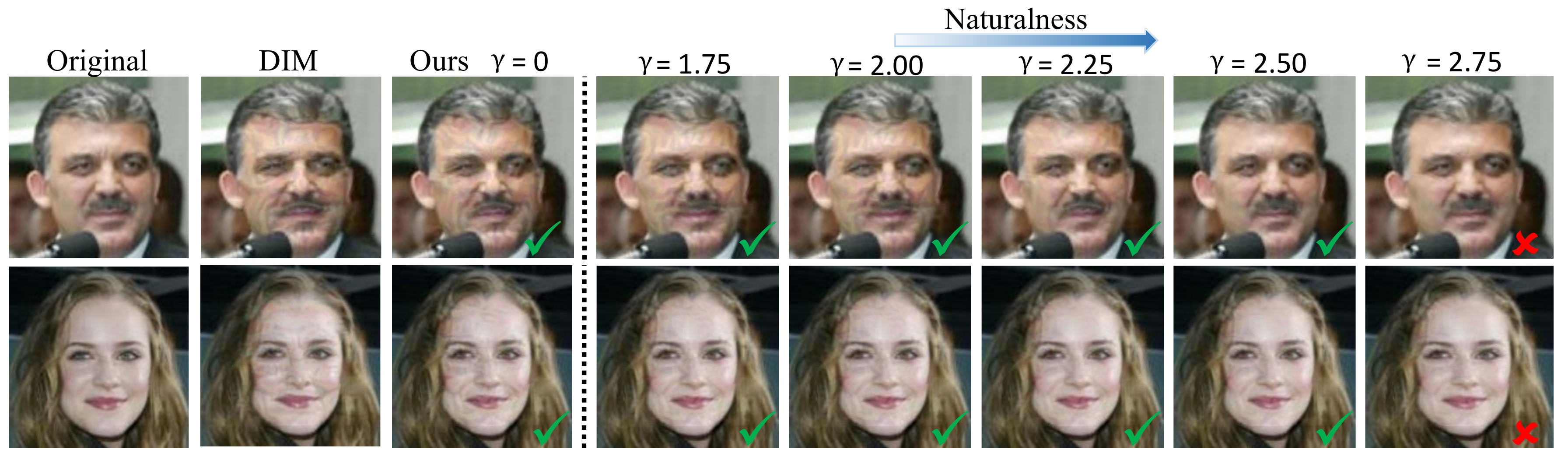}
\vspace{-1ex}
\caption{Experiments on how different  $\gamma$ affects the performance. Green hook refers to successful targeted identity protection while red hook refers to failure, which also implies a trade-off on effectiveness and naturalness. Best view when zoom in.}
\label{fig:natural}
\vspace{-3ex}
\end{figure*}

\begin{figure}[t]
\centering
\includegraphics[width=0.9\linewidth]{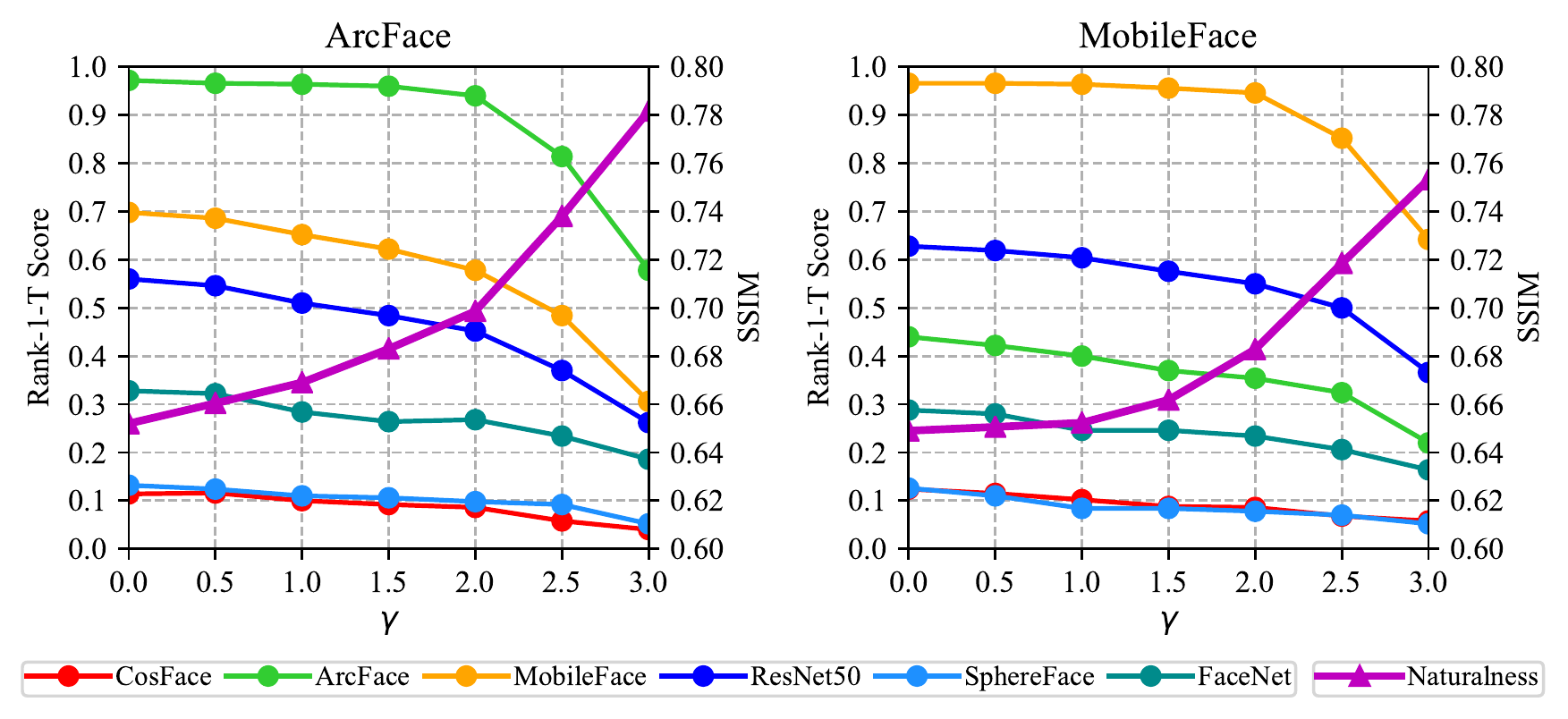}
\caption{Rank-1-T score and SSIM of protected images generated by TIP-IM with different $\gamma$ against different models.}
\label{fig:mmd}
\vspace{-3ex}
\end{figure}
\textbf{Comparison experiments about target images.} We test the performance of different numbers of targets in Appendix {\color{red} C}.  We experimentally find 10 target identities in this paper is enough, which implies that small increases in the number of targets can obtain impressive performance in spite of taking slight timing cost.  We specify some \emph{generated} images from StyleGAN~\cite{karras2020analyzing} as target images. The results show that our algorithm still has excellent black-box performance of identity protection. In practical applications, we can arbitrarily specify the available and authorized target identity set or generated face images, and our algorithm is applicable to any target set. 

\vspace{-0.1cm}
\subsection{Naturalness}
\vspace{-0.1cm}
To examine whether our algorithm is able to control the naturalness of protected images in the process of generating the samples, we perform experiments with different coefficient $\gamma$. Tab.~\ref{tab:nat} shows the evaluation results of different face recognition models (including ArcFace, MobileFace, and ResNet50) w.r.t three different metrics---PSNR, SSIM, and MMD. As $\gamma$ increases, the visual quality of the generated images is getting better based on different metrics, which is also consistent with the example in Fig.~\ref{fig:natural}. Therefore, 
conditioned on different coefficient $\gamma$, we can control the degree of the generated protected images. 
Apart from quantitative measures, we also performed naturalness manipulation for different $\gamma$ dynamically in Fig.~\ref{fig:natural}. The image looks more natural as the $\gamma$ increases, whereas to a certain extent identity protection tends to fail. We also perform a more general evaluation on all given recognition models in Fig.~\ref{fig:mmd}. As $\gamma$ increases, SSIM values perform a general downward trend for Rank-1-T accuracy, meaning that appropriate $\gamma$ is crucial for transferability and naturalness. 

\textbf{Practicability.} Face encryption focuses on generating effective and natural adversarial identity masks, which cannot be realized by most previous adversarial attacks w.r.t. the effectiveness (in Tab.~\ref{tab:transfer}) and naturalness (in Fig.~\ref{fig:ssim} and Fig.~\ref{fig:natural}). In practical applications, users can adopt proposed TIP-IM to adjust $\gamma$ to control stronger obfuscation  performance (effectiveness) or visual  quality (naturalness).

\begin{figure}[t]
\centering
\includegraphics[width=1.0\linewidth]{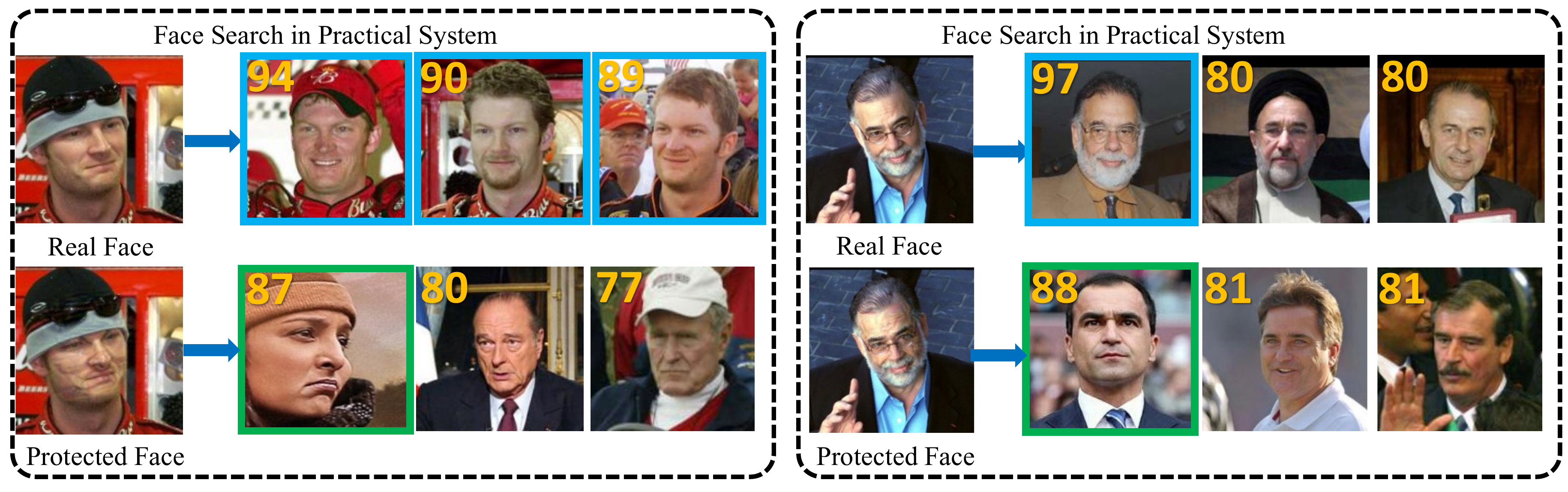}
\vspace{-1ex}
\caption{Examples of face encryption on the real-world face recognition API. We separately use real and protected faces by TIP-IM as probes to do face search and show top three results by similarity. Blue boxes represent the faces with same identities as probe faces and green boxes imply the faces belonging to targeted identities. Similarity scores with probe face are marked in yellow.}
\label{fig:tencent}
\vspace{-1ex}
\end{figure}

\vspace{-0.2cm}
\subsection{Effectiveness on a Real-World Application}
\vspace{-0.1cm}

In this section, we apply our proposed TIP-IM to test the identity protection performance on a commercial face search API available at Tencent AI Open Platform\footnote{\url{https://ai.qq.com/product/face.shtml}}. The working mechanism and training data are completely unknown for us. To simulate the privacy data scenario, we use the same gallery set described above. We choose $20$ probe faces from above probe set to execute face search based on similarity ranking in this platform. All $20$ probe faces can be identified at Rank1. Then we generate corresponding protected images from probe faces to execute face search. For return rankings there exists $6$ target identities in rank1 and $16$ in rank5. Note that those faces with the same identity also show a decreasing similarity in different degrees, which also illustrates the effectiveness for black-box face system, and two examples are shown in Fig.~\ref{fig:tencent}.


\section{Conclusion}
\vspace{-0.1cm}
In this paper, we studied the problem of identity protection by simulating realistic identification systems in the social media. Extensive experiments show that proposed TIP-IM method enables users to protect their private information from being exposed by the unauthorized identification systems while not affecting the user experience in social media. 

\section*{Acknowledgements}

{
This work was supported by the National Key Research and Development Program of China (No.s 2020AAA0104304, 2020AAA0106302), NSFC Projects (Nos. 61620106010, 62076147, U19A2081, U19B2034, U1811461),
Beijing Academy of Artificial Intelligence (BAAI), 
Alibaba Group through Alibaba Innovative Research Program, Tsinghua-Huawei Joint Research Program, a
grant from Tsinghua Institute for Guo Qiang, Tiangong Institute for Intelligent Computing, and the
NVIDIA NVAIL Program with GPU/DGX Acceleration.
}

\clearpage

\appendix

\section{Evaluation Results on MegFace.}

We report the results on the MegFace dataset in Tab.~\ref{tab:meg}. Compared with LFW and MegFace has more gallery images over 50k+ images. This large-scale challenging dataset results in more difficult targeted identity protection on the whole.

\section{Batch Analysis on MMD Optimization.}
Tab.~\ref{tab:bat} shows results on different batch sizes w.r.t naturalness. It can be seen that the evaluation of visual quality becomes stable as the batch size exceeds 50. We set batch size as 50 in this paper. For single image crafting, we have two choices. The first one is self augmentation including rotation, projective, brightness and transformations; The second one is collecting some irrelevant images to form a batch just for optimal results in the phase of MMD optimization. 

\begin{table}[!htp]
\scriptsize
\setlength{\tabcolsep}{2pt}
    \begin{center}
        
    \begin{tabular}{c|c|c|c|c|c|c|c|c|c}
    \hline
    
        & 10 & 20 & 30 & 40 & 50 & 60 & 70 & 80 & 90 \\
        \hline
        SSIM & 0.8518 & 0.8392 & 0.7592 & 0.7021& 0.6759 &0.6765 &0.6633 &0.6649 & 0.6674\\
        PSNR &28.71 &28.55 & 26.95& 26.02& 25.55& 25.63&25.40 &25.50& 25.54 \\
         \hline
         
    \end{tabular}
    \end{center}
    \caption{The mean PSNR (db) and SSIM for different batch sizes based on CosFace.}
    \label{tab:bat}
\end{table}


\section{Comparison Experiments about Target Images.}
\begin{figure}[!htp]
\centering
\includegraphics[width=0.45\linewidth]{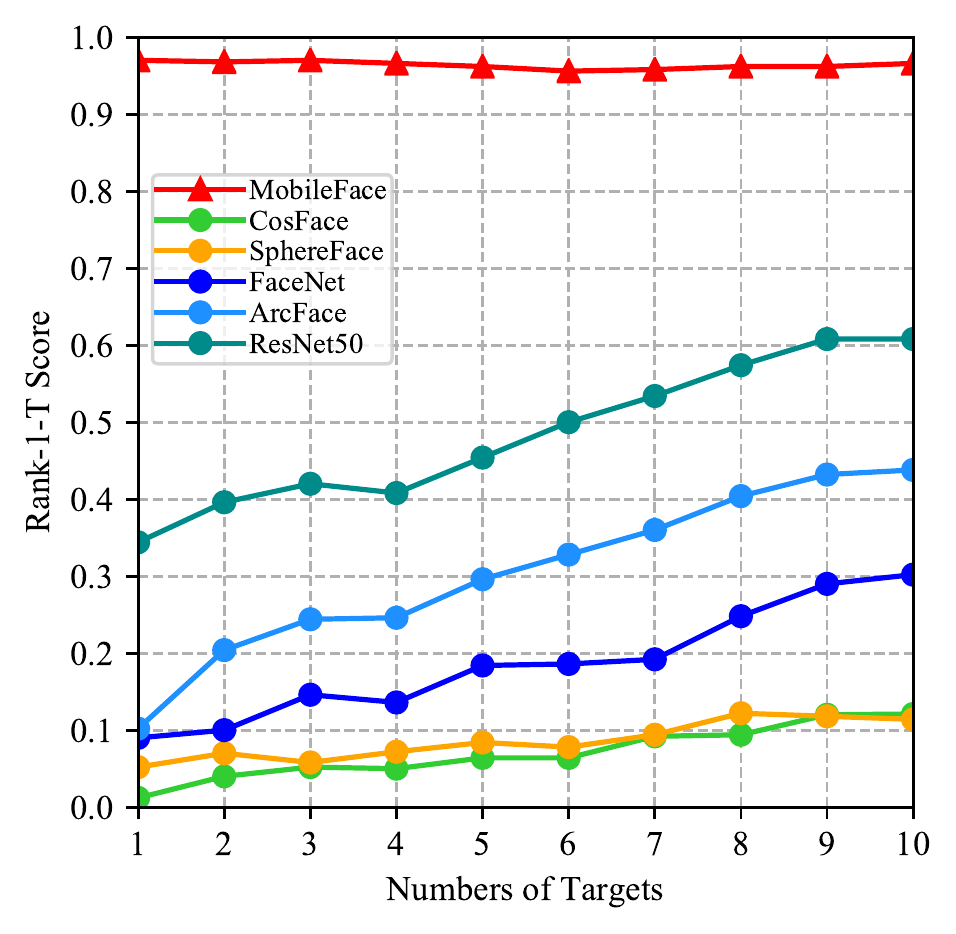}
\caption{The perturbation vs. numbers of targets curve of face identification models against black-box identity protection. \textit{MobileFace} is a surrogate white-box model.}
\label{fig:targets}
\vspace{-1ex}
\end{figure}

\textbf{Different numbers of targets.} As illustrated in Fig.~\ref{fig:targets}, we study the effect of different numbers on the black-box identity protection. The curve first rises and finally approaches the steady. Therefore, appropriate increases in the number of targets is beneficial to performance improvement against black-box models.

\begin{figure}[!htp]
\centering
\includegraphics[width=0.95\linewidth]{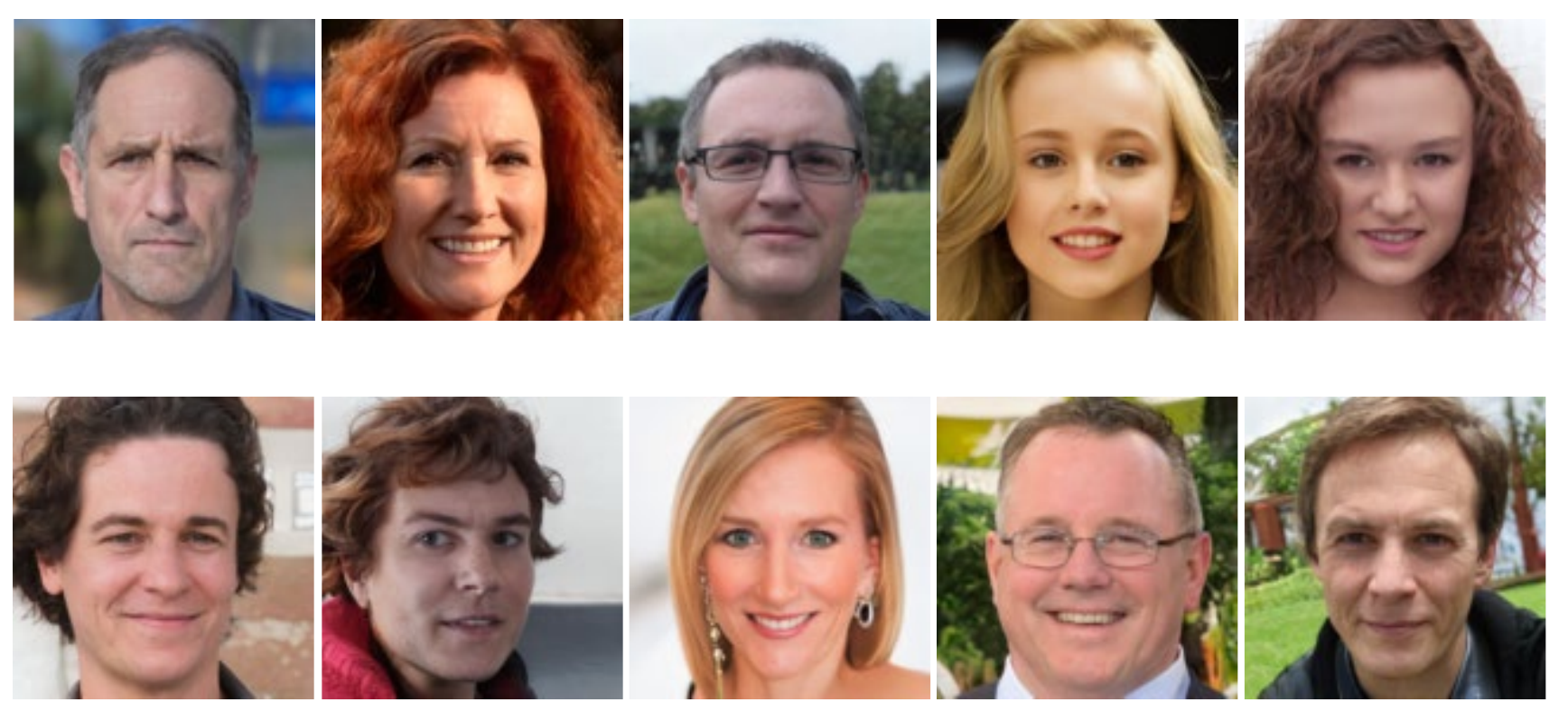}
\caption{Examples of some generated images from StyleGAN.}
\label{fig:gene}
\vspace{-1ex}
\end{figure}

\textbf{Generated images as targets.} To further verify that our algorithm does not depend on the selection of targets, we specify some generated images from StyleGAN~\cite{karras2020analyzing} as target images, which is illustrated as Fig.~\ref{fig:gene}. We use these generated images as target images and set the same other setting with above experiments. Tab.~\ref{tab:gene} shows Rank-1-T, Rand-1-UT, Rank-5-T and  Rank-5-UT  of  black-box  attacks  against CosFace, SphereFace, FaceNet, ArcFace, MobileFace and ResNet. The results show that our algorithm still has excellent black-box performance of identity protection. In practical
applications, we can arbitrarily specify the available and authorised target identity set or some generated facial images, and our algorithm is applicable to any target set. 

\begin{table}[!htp]
\scriptsize
\setlength{\tabcolsep}{3pt}
    \begin{center}
        
    \begin{tabular}{c|c|c|c|c|c|c}
    \hline
    
        & ArcFace & MobileFace & ResNet50 & SphereFace & FaceNet & CosFace  \\
        \hline
        Rank-1-T &12.2 &32.4 &29.2 &27.4 &28.2 & 49.6 \\
        Rank-5-T &30.4 &52.2 &54.8 &56.4 & 53.8 &70.0\\
        Rank-1-UT &89.0 &73.8 &49.6 &60.8 &54.2 & 95.0\\
        Rank-5-UT &82.0 &55.8 &31.6 &41.8 &34.8 & 93.6\\
        
         \hline
         
    \end{tabular}
    \end{center}
    \caption{Results of  black-box  attacks  against SphereFace, FaceNet, ArcFace, MobileFace, ResNet and CosFace when treating the \emph{generated} images as the target images.}
    \label{tab:gene}
\end{table}

\begin{figure*}[t]
\centering
\includegraphics[width=0.99\linewidth]{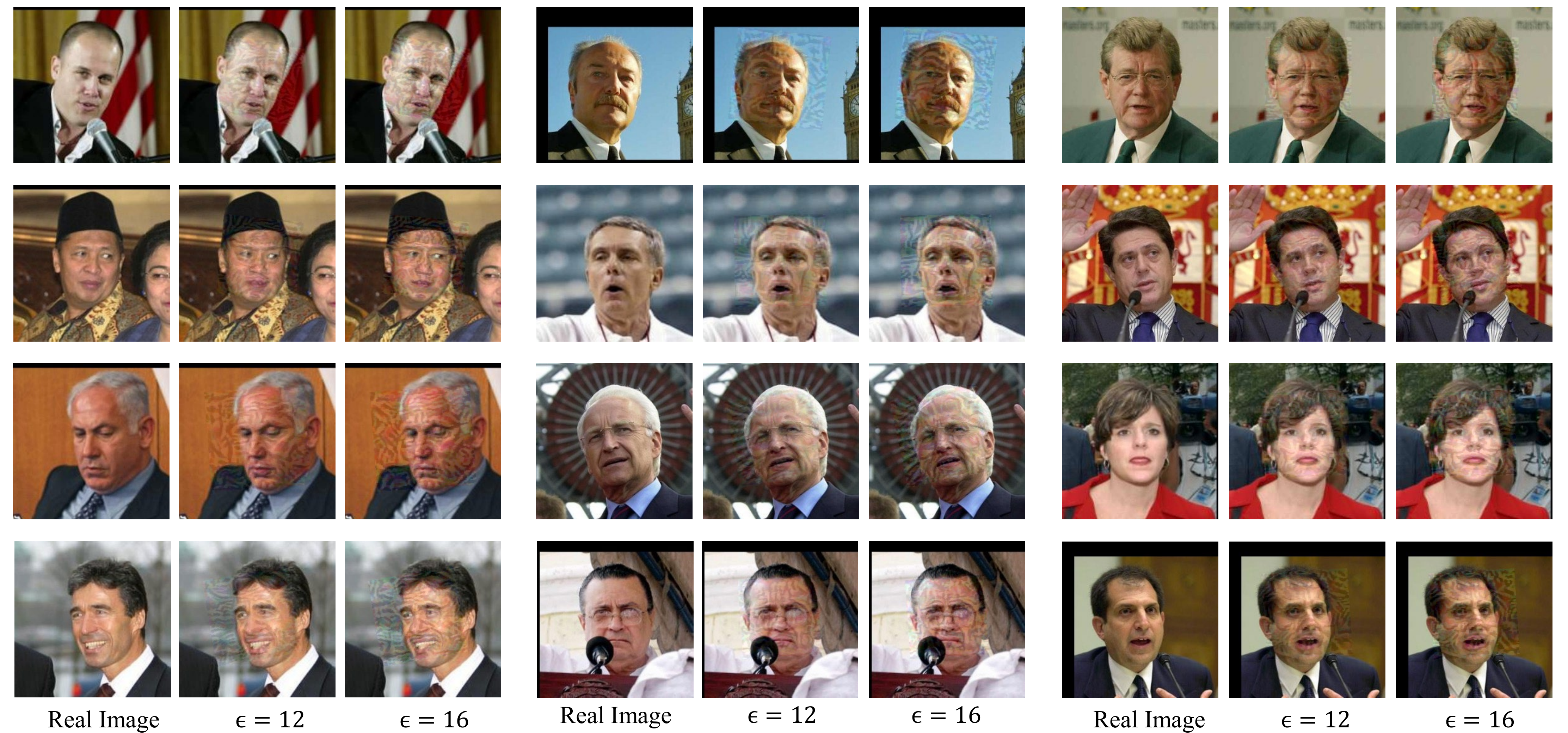}
\caption{More examples for different perturbations under the $l_\infty$ norm by existing adversarial methods.}
\label{fig:more}

\end{figure*}

\begin{table*}[t]
    \begin{center}
    \scriptsize
    \setlength{\tabcolsep}{9pt}
    \begin{tabular}{c|c|cc|cc|cc|cc|cc|cc}
    \hline
        \multirow{2}{*}{} & \multirow{2}{*}{Attack} & 
        \multicolumn{2}{c|}{ArcFace} & \multicolumn{2}{c|}{MobileFace} & \multicolumn{2}{c|}{ResNet50} & \multicolumn{2}{c|}{SphereFace} & \multicolumn{2}{c|}{FaceNet} & \multicolumn{2}{c}{CosFace} \\
         \cline{3-14}
         & & R1-T & R5-T & R1-T & R5-T & R1-T & R5-T & R1-T & R5-T & R1-T & R5-T & R1-T & R5-T \\
         \hline
         
         ArcFace& \tabincell{c}{DIM~\cite{xie2019improving} \\ TIP-IM} & 
         \tabincell{c}{92.0$^*$ \\97.2$^*$} & 
         \tabincell{c}{97.0$^*$ \\98.4$^*$} & 
         \tabincell{c}{11.6 \\60.1} & 
         \tabincell{c}{36.0 \\81.4} & 
         \tabincell{c}{8.0 \\51.4} & 
         \tabincell{c}{23.4 \\67.4} &
         \tabincell{c}{1.6 \\10.1} & 
         \tabincell{c}{7.6 \\19.5} & 
         \tabincell{c}{3.4 \\25.3} & 
         \tabincell{c}{13.4 \\43.4} &
         \tabincell{c}{1.4 \\11.1} & 
         \tabincell{c}{7.6 \\24.2}\\
         \hline
         MobileFace& \tabincell{c}{DIM~\cite{xie2019improving} \\ TIP-IM} & 
         \tabincell{c}{5.8 \\37.4} & 
         \tabincell{c}{19.4 \\57.2} & 
         \tabincell{c}{94.8$^*$ \\97.0$^*$} & 
         \tabincell{c}{96.4$^*$ \\97.2$^*$} & 
         \tabincell{c}{18.6 \\52.9} & 
         \tabincell{c}{42.0 \\73.5} &
         \tabincell{c}{2.8 \\9.8} & 
         \tabincell{c}{9.6 \\19.8} & 
         \tabincell{c}{4.0 \\20.9} & 
         \tabincell{c}{14.2 \\36.2} &
         \tabincell{c}{2.0 \\10.5} & 
         \tabincell{c}{9.2 \\22.1}\\
         \hline
         ResNet50& \tabincell{c}{DIM~\cite{xie2019improving} \\ TIP-IM} &
         \tabincell{c}{7.8 \\31.1} & 
         \tabincell{c}{19.0 \\45.2} & 
         \tabincell{c}{15.2 \\56.9} & 
         \tabincell{c}{44.2 \\76.3} & 
         \tabincell{c}{91.4$^*$ \\92.1$^*$} & 
         \tabincell{c}{95.4$^*$ \\97.5$^*$} &
         \tabincell{c}{2.4 \\11.6} & 
         \tabincell{c}{13.0 \\21.2} & 
         \tabincell{c}{4.8 \\20.5} & 
         \tabincell{c}{16.8 \\35.2} &
          \tabincell{c}{3.2 \\10.0} & 
         \tabincell{c}{9.2 \\25.2}\\

         \hline
    \end{tabular}
    \end{center}
    \caption{Rank-1-T and Rank-5-T (\%) of black-box attacks against CosFace, SphereFace, FaceNet, ArcFace, MobileFace and ResNet on MegFace.  $^*$ indicates white-box attacks.}
    \label{tab:meg}
\end{table*}

\section{Ill-suited $\ell_{p}$-norm perturbation in Face encryption}

Face encryption focuses on generating adversarial identity masks that can be overlaid on facial images without sacrificing the visual quality. As illustrated in Fig.~\ref{fig:more}, although the adversarial perturbations generated by the existing attack methods, \eg, PGD and MIM, have a small intensity change (e.g., $12$ or $16$ for each pixel in $[0,255]$), they may still sacrifice the visual quality for human perception due to the artifacts. $\ell_{p}$-norm adversarial perturbations can not naturally fit human perception well, which also accords with~\cite{zhao2017generating,sen2019should}. Thus proposed TIP-IM introduces a better multi-target optimization mechanism to improve effectiveness and $\mathcal{L}_{nat}$ in the objective of Eq.~(2) to generate more natural images.

{\small
\bibliographystyle{ieee_fullname}
\bibliography{egbib}
}
\end{document}